\documentclass{article}
\usepackage{ijcai22}

\usepackage{times}
\usepackage{soul}
\usepackage{url}
\usepackage[hidelinks]{hyperref}
\usepackage[utf8]{inputenc}
\usepackage[small]{caption}
\usepackage{graphicx}
\usepackage{amsmath}
\usepackage{amsthm}
\usepackage{booktabs}
\usepackage{algorithm}
\usepackage{algorithmic}
\usepackage{amssymb}
\usepackage{subfigure}
\usepackage{multirow}

\newcommand{\secref}[1]{Section~\ref{#1}}
\newcommand{\figref}[1]{Figure~\ref{#1}}
\newcommand{\tabref}[1]{Table~\ref{#1}}
\newcommand{\algref}[1]{Algorithm~\ref{#1}}
\newcommand{\eqnref}[1]{Eq.~(\ref{#1})}

\newcommand{\argmax}{\operatornamewithlimits{\arg \max}}

\newcommand{\vectorize}{v}
\newcommand{\DP}{\textrm{DP}}
\newcommand{\MH}{\textrm{MH}}
\newcommand{\frnsel}{\textrm{FAN-Select}}
\newcommand{\frnloc}{\textrm{FAN-Pose}}

\newcommand{\bstau}{\boldsymbol \tau}

\newcommand{\bh}{\mathbf{h}}

\newcommand{\bt}{\mathbf{t}}

\newcommand{\bx}{\mathbf{x}}
\newcommand{\by}{\mathbf{y}}

\newcommand{\br}{\mathbf{r}}

\newcommand{\bc}{\mathbf{c}}

\newcommand{\bW}{\mathbf{W}}
\newcommand{\bX}{\mathbf{X}}
\newcommand{\bY}{\mathbf{Y}}
\newcommand{\bM}{\mathbf{M}}
\newcommand{\bQ}{\mathbf{Q}}
\newcommand{\bK}{\mathbf{K}}
\newcommand{\bV}{\mathbf{V}}

\newcommand{\calH}{\mathcal{H}}
\newcommand{\calL}{\mathcal{L}}

\newcommand{\calX}{\mathcal{X}}

\newcommand{\bbR}{\mathbb{R}}

\newcommand{\shapesquare}{\texttt{Square}}
\newcommand{\shapemondrian}{\texttt{Mondrian-Square}}
\newcommand{\shapepentagon}{\texttt{Pentagon}}
\newcommand{\shapehexagon}{\texttt{Hexagon}}

\graphicspath{{./figures/}}

\title{Learning to Assemble Geometric Shapes}

\author{
Jinhwi Lee$^{1,2}$\thanks{Equal contribution.}
\and
Jungtaek Kim$^1$\footnotemark[1]
\and
Hyunsoo Chung$^1$
\and
Jaesik Park$^1$
\And
Minsu Cho$^1$
\affiliations
$^1$Pohang University of Science and Technology (POSTECH)\\
$^2$POSCO\\
\emails
\{jinhwi, jtkim, hschung2, jaesik.park, mscho\}@postech.ac.kr
}

\begin{document}

\maketitle

\begin{abstract}
Assembling parts into an object is a combinatorial problem that arises 
in a variety of contexts in the real world and involves numerous applications 
in science and engineering.
Previous related work tackles limited cases with identical unit parts 
or jigsaw-style parts of textured shapes, which greatly mitigate 
combinatorial challenges of the problem.
In this work, we introduce the more challenging problem of shape assembly, which involves textureless fragments of arbitrary shapes with indistinctive junctions, 
and then propose a learning-based approach to solving it.
We demonstrate the effectiveness on shape assembly tasks 
with various scenarios, including 
the ones with abnormal fragments (e.g., missing and distorted), the different number of fragments, 
and different rotation discretization.
\end{abstract}

% Introduction
% !TEX root =  ../main.tex
\section{Introduction\label{sec:intro}}

We humans show an excellent ability in solving a shape assembly problem, e.g., tangram~\cite{SlocumJ2003book},
by analyzing a target object and its constituent parts.\footnote{Supplementary material and implementations are available at \href{https://github.com/POSTECH-CVLab/LAGS}{https://github.com/POSTECH-CVLab/LAGS}.}
Machines, however, still fall short of the level of intelligence and often suffer from the combinatorial nature of assembly problems; the number of possible configurations drastically increases with the number of parts. Developing an effective learner that tackles the issue is crucial since the assembly problems are pervasive in science and engineering fields
such as manufacturing processes, structure construction, and real-world robot automation~\cite{ZakkaK2020icra}.

There have been a large volume of studies~\cite{LodiA2002ejor} 
on different types of shape assembly or composition problems 
in a variety of fields such as 
biology~\cite{SanchesCAA2009amc}, earth science~\cite{TorsvikTH2003science}, 
archaeology~\cite{DerechN2018arxiv}, and tiling puzzles~\cite{NorooziM2016eccv}.
When part shapes have neither distinct textures nor mating junctions, 
the assembly problem can be formulated as a \emph{combinatorial optimization} problem, 
which aims to occupy a target object 
using candidate parts~\cite{CoffmanEG1996survey}, 
such as tangram~\cite{SlocumJ2003book}.
It is also related to the classic problem of bin packing~\cite{BrownAR1971book,GoyalA2020icml}, 
which is one of the representative problems in combinatorial optimization.
Most previous related work, however, tackles shape assembly problems by exploiting mating patterns across parts, e.g., compatible junctions or textures of candidate parts.
For example, a jigsaw puzzle problem is typically solved by comparing and matching patterns across fragments where each fragment contains a distinct pattern~\cite{ChoTS2010cvpr}.
In general cases where such hints are unavailable (i.e., \emph{textureless parts} and \emph{indistinctive junctions}), 
assembling fragments becomes significantly more challenging.

\begin{figure}
	\centering
	\includegraphics[width=0.115\textwidth]{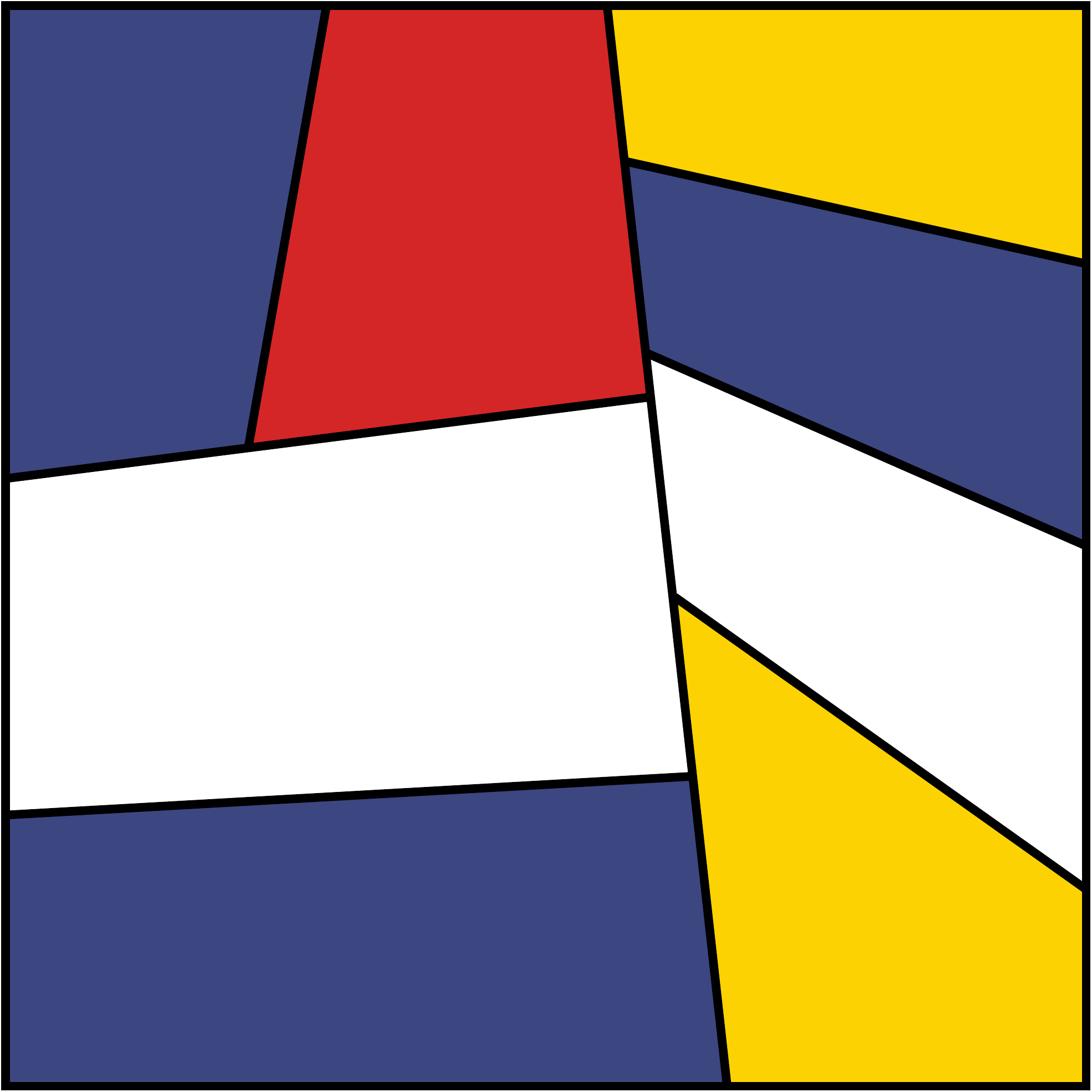}
	\includegraphics[width=0.115\textwidth]{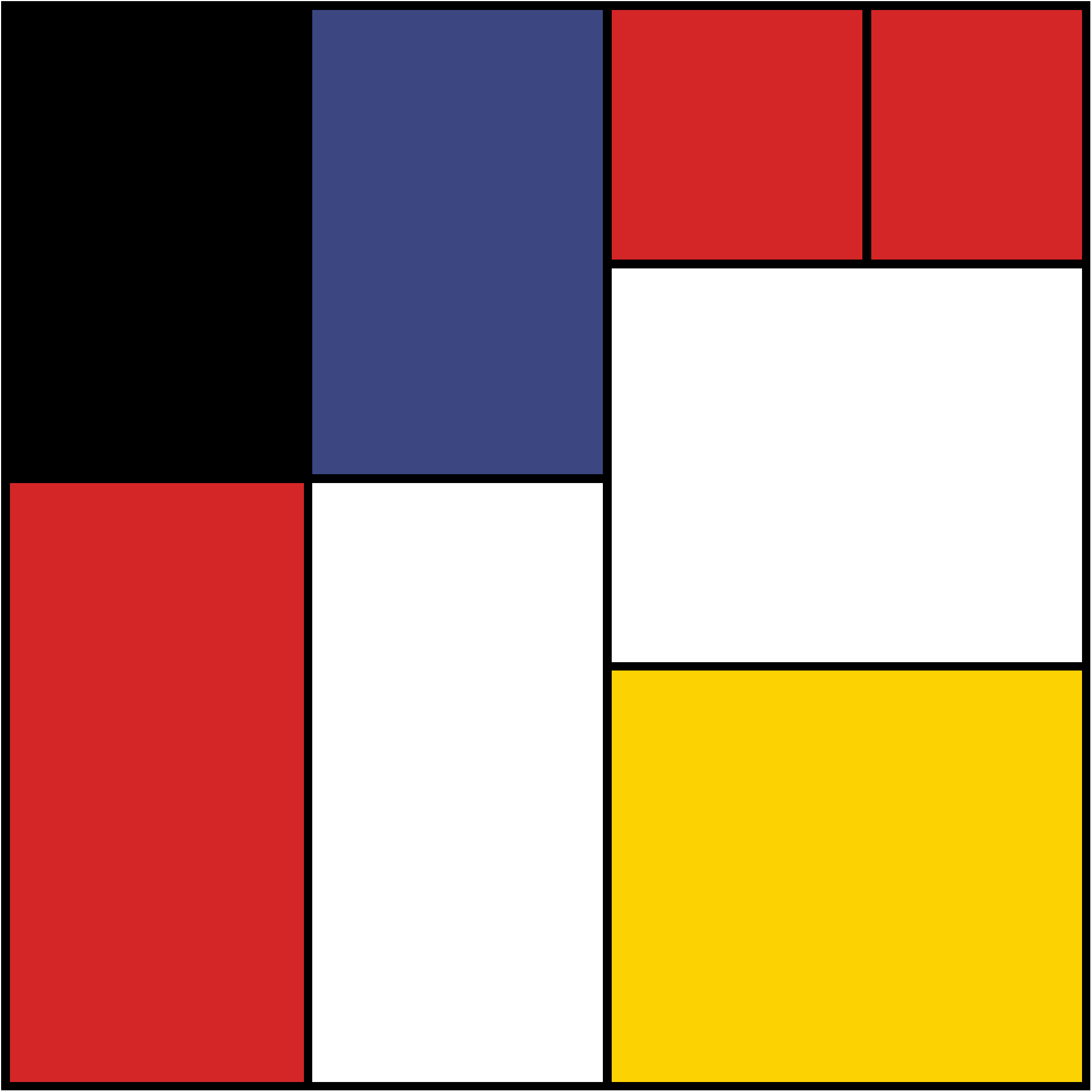}
	\includegraphics[width=0.115\textwidth]{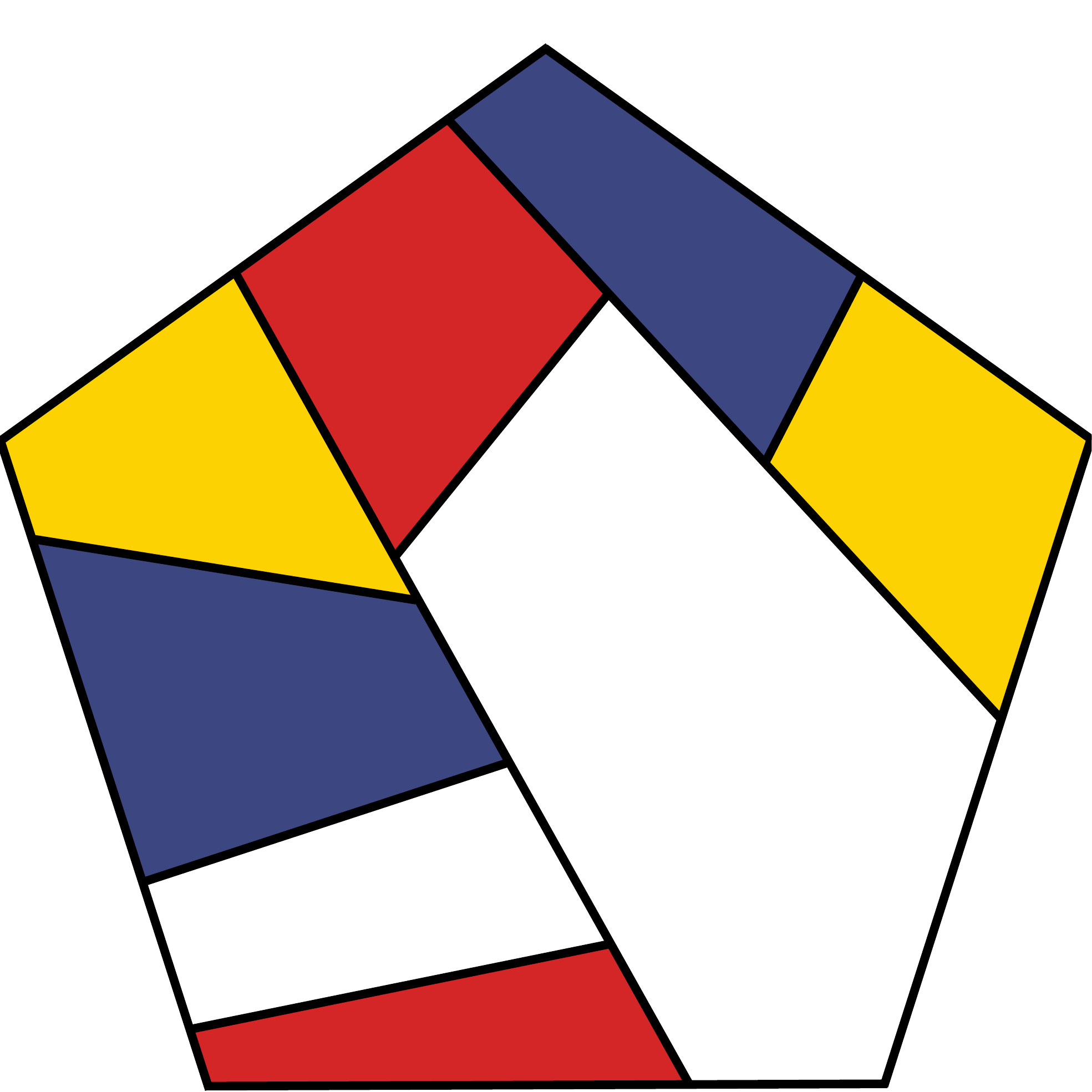}
	\includegraphics[width=0.115\textwidth]{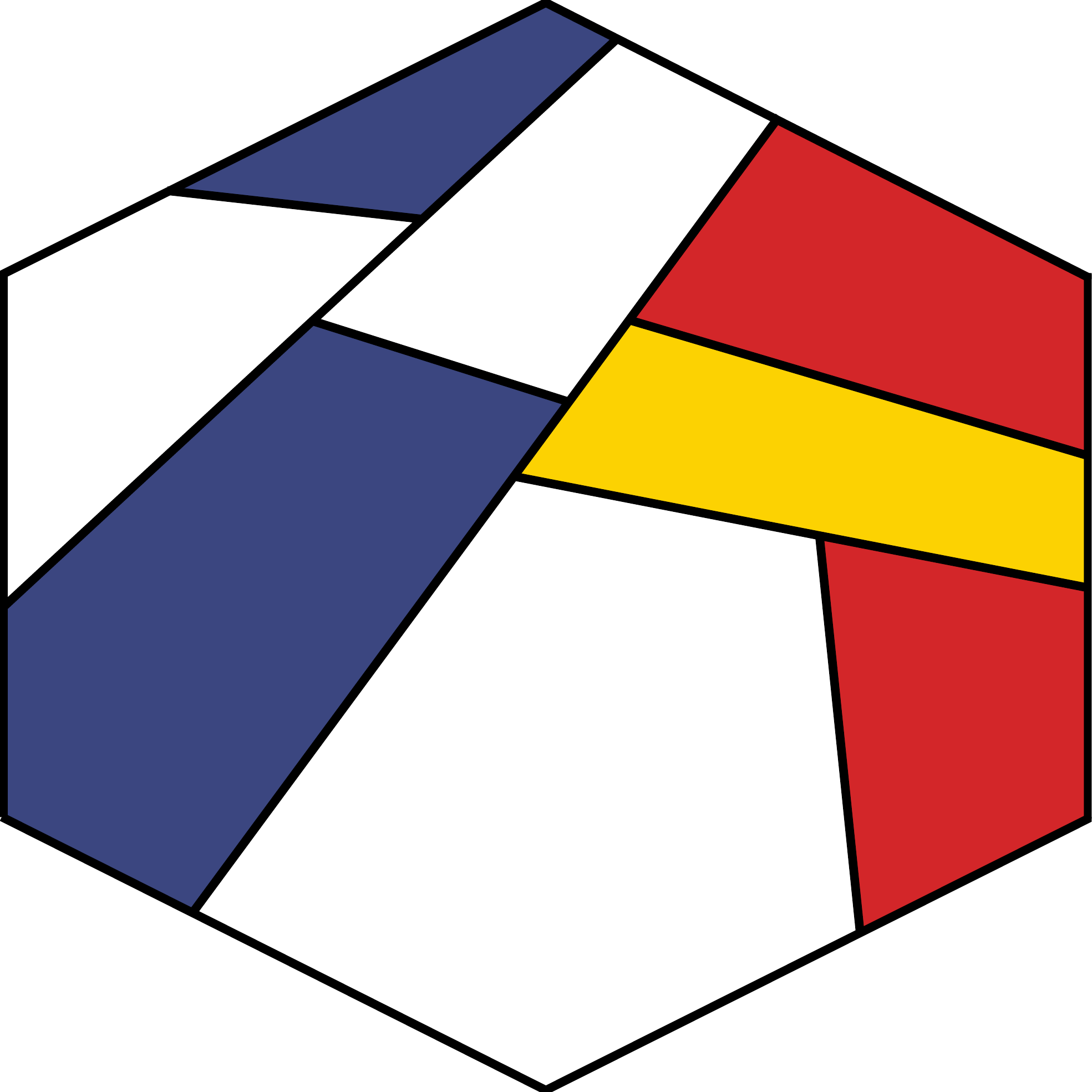}
	\caption{Geometric shape assembly.
		Given randomly-partitioned fragments, it is challenging to assemble them into the target shape.
		\label{fig:partition}}
\end{figure}

In this work we first introduce a simple yet challenging problem of assembly and then propose an efficient and effective learning-based approach to solving this problem. 
We split a two-dimensional target shape into multiple fragments of arbitrary polygons by a stochastic partitioning process~\cite{SchumacherR1969techreport,RoyDM2008neurips} and ask an agent to assemble the target shape given the partitioned fragments while the original poses are hidden.
Given a target object and a set of candidate fragments, 
the proposed model learns to select one of the fragments 
and place it into a right place. It is designed to process the candidate fragments in a permutation-equivariant manner and is capable of generalizing to cases with an arbitrary number of fragments.

Our experiments show that the proposed method effectively learns to tackle different assembly scenarios such as those with abnormal fragments, 
different rotation discretization, and colored textures, whereas a brute-force search, metaheuristic optimization~\cite{BoussaidI2013is}, 
and Bayesian optimization approach~\cite{BrochuE2010arxiv} easily fall into a bad local optimum.
% Partitioning
% !TEX root =  ../main.tex

\section{Problem Definition\label{sec:problem}}

Suppose that 
we are given textureless fragments with indistinctive junctions.
Our goal is to sequentially assemble the given fragments, 
which is analogous to a tangram puzzle. 
Unlike an approach based on the backtracking method, 
which considers how remaining fragments will fit into unfilled area of the target shape, 
we attempt to solve such a problem with a learning-based method.

\paragraph{Geometric Shape Assembly.}

Suppose that we are given a set of fragments $\calX = \{ \bx_i \}_{i=1}^{N}$ and a target geometric shape $S$ on a space $\Phi$.
In this work, we assume that both fragments and a target shape are defined on a two-dimensional space. 
We sequentially assemble those fragments into the target shape; 
at each step, a fragment $\bx_i$ is sampled without replacement 
and placed on top of a current shape $\bc$.
Our goal is to reconstruct $S$ consuming all the fragments.

\paragraph{Shape Fragmentation Dataset.}

We create a dataset by partitioning a shape into multiple fragments, 
which can be easily used to pose its inverse task, i.e., an assembly problem.
Inspired by binary space partitioning algorithm~\cite{SchumacherR1969techreport}, 
we randomly split a target shape, create a set of random fragments for each target shape, 
choose the order of fragments, and rotate them at random.
In particular, possible rotation is discretized to a fixed number of bins.
After $K$ times of binary partitioning, 
we obtain $N = 2^K$ fragments.
The details are described in the supplementary material.

\begin{figure*}[t]
	\centering
	\includegraphics[width=0.95\textwidth]{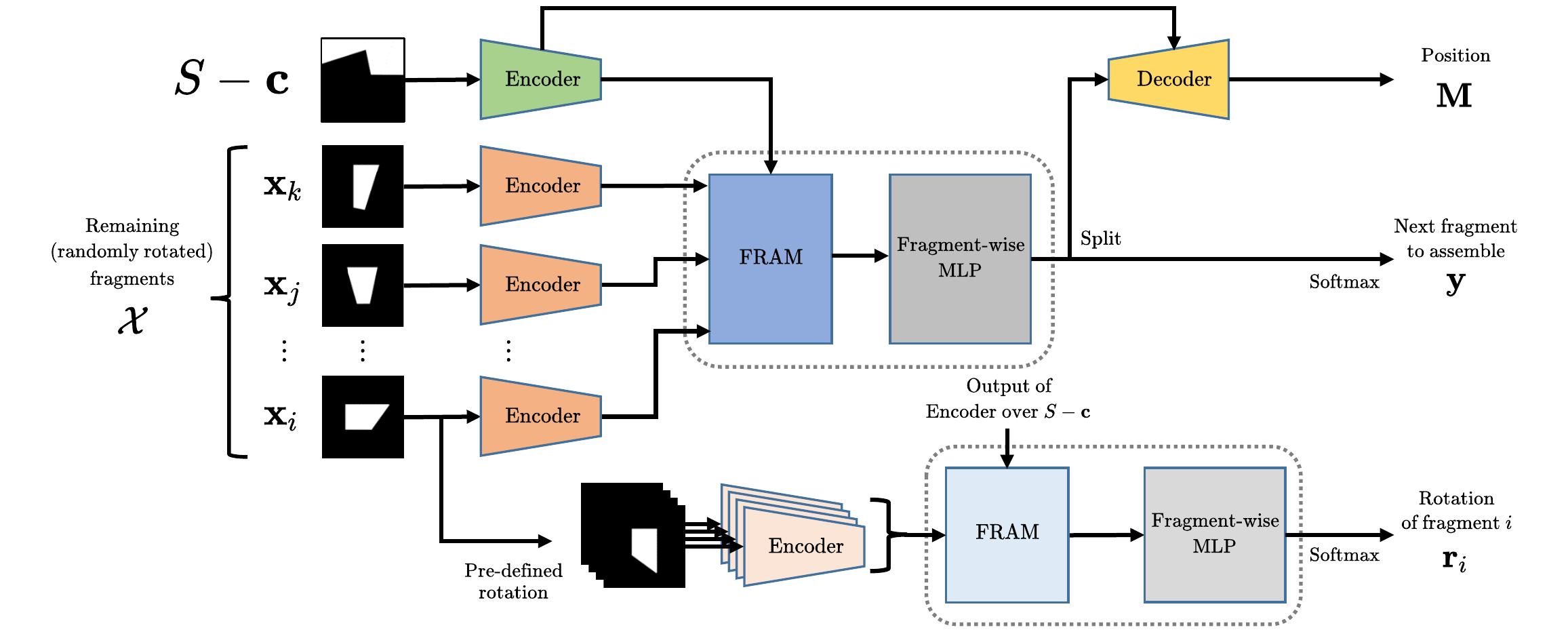}
	\caption{Overall architecture of the fragment assembly network, which includes all the components, e.g., a rotation branch, in~\secref{sec:exp}.\label{fig:overall}}
\end{figure*}

\section{Geometric Shape Assembly\label{sec:method}}

To tackle the assembly problem, we propose a model that learns to select a fragment from candidates and place it on top of a current shape.
The model is fed by two inputs:
\begin{enumerate}
    \item remaining shape $S - \bc$, which is a shape to assemble with the fragments excluding the fragments already placed;
    \item a set of fragments $\calX$, which consists of all candidates for the next assembly.
\end{enumerate}
It then predicts which $\bx_i \in \calX$ should be used and where it should be placed on $\Phi$.
We assemble all the fragments by iteratively running our model until no candidate remains.

To carefully make the next decision, 
the model needs to extract geometric information 
of the fragments and the target object, and also understand their relations.
Our model is trained through the supervision obtained from 
shape fragmentation processes, which is expressed with a large number of episodes.
Each episode contains the information about which fragment is the next 
and where the fragment is placed.

\subsection{Fragment Assembly Networks}

Our model, dubbed Fragment Assembly Network (FAN), considers a candidate fragment $\bx_i$ in the context of the other candidates $\calX \setminus \bx_i$ and the remaining shape $S - \bc$, and produces three outputs:    
(i) selection probability for $\bx_i$, which implies how likely $\bx_i$ is selected; 
(ii) placement probability map for $\bx_i$, which means how likely each position is for $\bx_i$;
(optional, iii) rotation probability for $\bx_i$, which indicates a bin index with respect to rotation angles.
As shown in~\figref{fig:overall}, FAN contains two branches for the outputs, fragment selection network (i.e., $\frnsel$) and fragment placement network (i.e., $\frnloc$).
Both networks share many learnable parameters 
that are parts of encoders, Fragment Relation Attention Module (FRAM) and fragment-wise MLP.
FRAM, inspired by a Transformer network~\cite{VaswaniA2017neurips}, captures the relational information between the fragments.

\paragraph{Fragment Selection.}
$\frnsel$ is a binary classifier taking as inputs $\bx_i$, $S - \bc$, and $\calX\setminus\bx_i$:
\begin{equation}
    y_i = \frnsel\left(\bx_i; S - \bc, \calX\setminus\bx_i \right) \in \bbR,
    \label{eqn:frn_select}
\end{equation}
where $\bx_i \in \calX$.
Since, in particular, \eqnref{eqn:frn_select} takes into account 
relationship between fragments with FRAM, 
it can choose the next fragment by considering the fragment candidates and remaining shape.
Furthermore, the number of parameters in FAN does not depend on 
the number of fragments, which implies that 
the cardinality of $\calX$ can be varied.
This property helps our network to handle 
variable-length set of fragments.
Note that $\by = [y_i, \ldots, y_{|\calX|}] \in \bbR^{|\calX|}$.

\paragraph{Fragment Placement.}
$\frnloc$ determines the pose of $\bx \in \calX$ by predicting a probability map over pixels and a probability with respect to rotation angle bins:
\begin{equation}
    \bM_i(, \br_i) = \frnloc \left( \bx_i; S - \bc, \calX\setminus\bx_i, \calX'_i \right),
    \label{eqn:frn_place}
\end{equation}
where $\calX'_i$ is a set of all possible pre-defined rotation of $\bx_i$.
Note that we simplify our problem with pre-defined discrete bins of rotation angles.
This network is implemented as an encoder-decoder architecture with skip connections between them, 
in order to compute pixel-wise probabilities with convolution and transpose convolution operations.
This structure reduces the number of parameters due to the absence of the last fully-connected layer.
Note that $\bM_i \in \bbR^{w \times h}$ and $\br_i \in \bbR^b$ where $b$ is the number of rotation angle bins.

\begin{figure}[t]
    \centering
    \subfigure[Simulated annealing]{
        \includegraphics[width=0.11\textwidth]{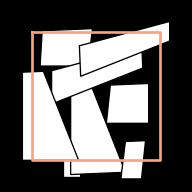}
        \includegraphics[width=0.11\textwidth]{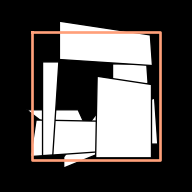}
        \includegraphics[width=0.11\textwidth]{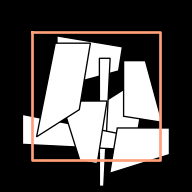}
        \includegraphics[width=0.11\textwidth]{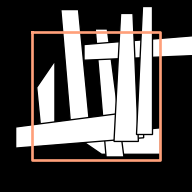}
    }
    \subfigure[Bayesian optimization]{
        \includegraphics[width=0.11\textwidth]{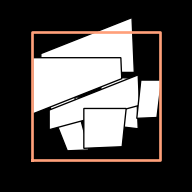}
        \includegraphics[width=0.11\textwidth]{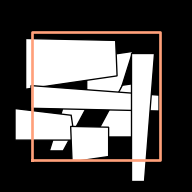}
        \includegraphics[width=0.11\textwidth]{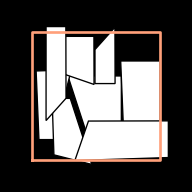}
        \includegraphics[width=0.11\textwidth]{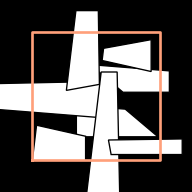}
    }
    \subfigure[V-GAN]{
        \includegraphics[width=0.11\textwidth]{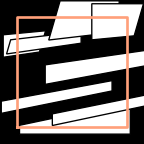}
        \includegraphics[width=0.11\textwidth]{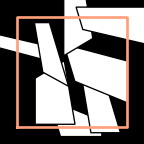}
        \includegraphics[width=0.11\textwidth]{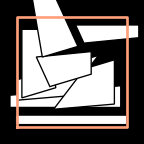}
        \includegraphics[width=0.11\textwidth]{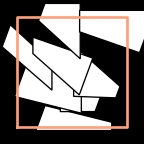}
    }
    \subfigure[Ours]{
        \includegraphics[width=0.11\textwidth]{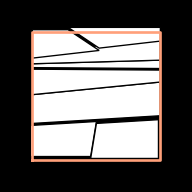}
        \includegraphics[width=0.11\textwidth]{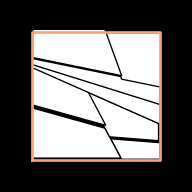}
        \includegraphics[width=0.11\textwidth]{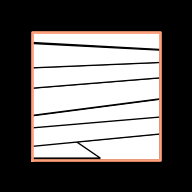}
        \includegraphics[width=0.11\textwidth]{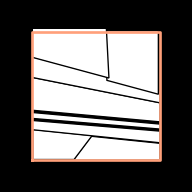}
    }
    \caption{Assembling results for \shapesquare.\label{fig:assembly_square}}
\end{figure}

\begin{figure}[t]
    \centering
    \subfigure[Simulated annealing]{
        \includegraphics[width=0.11\textwidth]{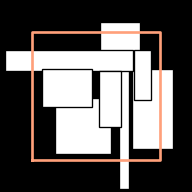}
        \includegraphics[width=0.11\textwidth]{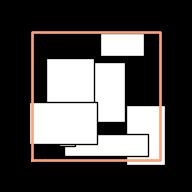}
        \includegraphics[width=0.11\textwidth]{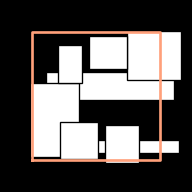}
        \includegraphics[width=0.11\textwidth]{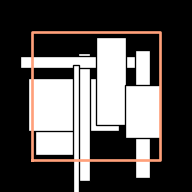}
    }
    \subfigure[Bayesian optimization]{
        \includegraphics[width=0.11\textwidth]{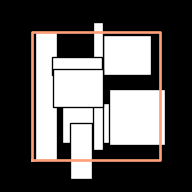}
        \includegraphics[width=0.11\textwidth]{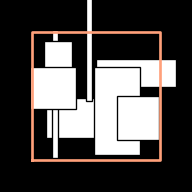}
        \includegraphics[width=0.11\textwidth]{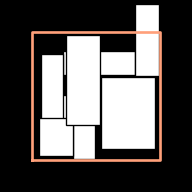}
        \includegraphics[width=0.11\textwidth]{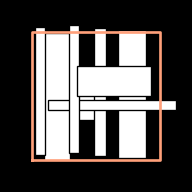}
    }
    \subfigure[V-GAN]{
        \includegraphics[width=0.11\textwidth]{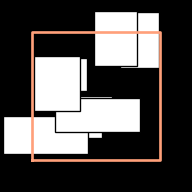}
        \includegraphics[width=0.11\textwidth]{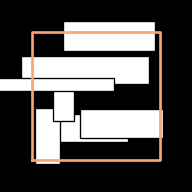}
        \includegraphics[width=0.11\textwidth]{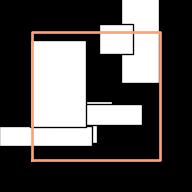}
        \includegraphics[width=0.11\textwidth]{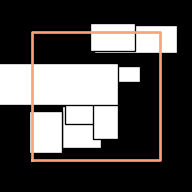}
    }
    \subfigure[Ours]{
        \includegraphics[width=0.11\textwidth]{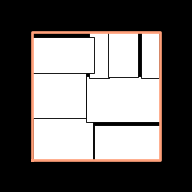}
        \includegraphics[width=0.11\textwidth]{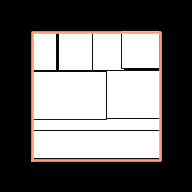}
        \includegraphics[width=0.11\textwidth]{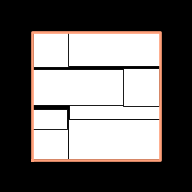}
        \includegraphics[width=0.11\textwidth]{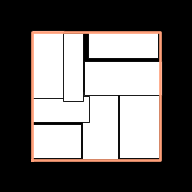}
    }
    \caption{Assembling results for \shapemondrian.\label{fig:assembly_mondrian}}
\end{figure}

\paragraph{Fragment Relation Attention Modules.}
We suggest attention-based FRAM, 
which considers high-order relations between the remaining fragments 
using multi-head attentions~\cite{VaswaniA2017neurips}.
%As proposed by \cite{VaswaniA2017neurips}, 
Given $\bX \in \bbR^{n_1 \times d}$ and $\bY \in \bbR^{n_2 \times d}$, 
multi-head attention is composed of multiple scaled dot-product attentions:
\begin{align}
	\DP(\bQ, \bK, \bV) &= \sigma \left( \frac{\bQ \bK^\top}{\sqrt{d_1}} \right) \bV,\\
	\MH(\bX, \bY, \bY) &= [\DP_1, \ldots, \DP_h] \bW^O,
\end{align}
where $\bQ \in \bbR^{n_1 \times d_1}$, $\bK \in \bbR^{n_2 \times d_1}$, 
$\bV \in \bbR^{n_2 \times d_2}$, 
$\DP_i = \DP(\bX \bW_i^{\bQ}, \bY \bW_i^{\bK}, \bY \bW_i^{\bV})$, 
$\sigma$ is a softmax function, 
$h$ is the number of heads, and $\bW^O \in \bbR^{h d_2 \times d}$ is the parameters.
To leverage the expression power of multi-head attention module, 
$\bX$ and $\bY$ are projected by the different parameter sets $\bW^\bQ \in \bbR^{d \times d_1}$, 
$\bW^\bK \in \bbR^{d \times d_1}$, and $\bW^\bV \in \bbR^{d \times d_2}$.

\begin{table}[t]
	\centering
	\scriptsize
	\setlength{\tabcolsep}{3pt}
    \begin{tabular}{ccccccc}
        \toprule
        \textbf{Target Shape} & \textbf{Method} & \textbf{Cov@0.95} & \textbf{Cov@0.90} & \textbf{Cov} & \textbf{IoU} & \textbf{Time (sec.)} \\
        \midrule
        \multirow{4}{*}{\shapesquare} & SA & 0.000 & 0.000 & 0.720 & 0.655 & 1,860\\
        & BayesOpt & 0.000 & 0.000 & 0.730 & 0.664 & 155\\
        & V-GAN & 0.000 & 0.000 & 0.720 & 0.562 & $\boldsymbol <$ \textbf{1}\\
        & Ours & \textbf{0.470} & \textbf{0.649} & \textbf{0.914} & \textbf{0.882} & $\boldsymbol <$ \textbf{1}\\
        \midrule
        \multirow{3}{*}{\shapepentagon} & SA & 0.000 & 0.000 & 0.697 & 0.581 & 1,854\\
        & BayesOpt & 0.000 & 0.000 & 0.710 & 0.660 & 149\\
%        V-GAN & \multicolumn{5}{c}{N/A} \\
        & Ours & \textbf{0.452} & \textbf{0.696} & \textbf{0.922} & \textbf{0.884} & $\boldsymbol <$ \textbf{1}\\
        \midrule
        \multirow{3}{*}{\shapehexagon} & SA & 0.000 & 0.000 & 0.711 & 0.608 & 1,846\\
        & BayesOpt & 0.000 & 0.000 & 0.727 & 0.674 & 153\\
%        V-GAN & \multicolumn{5}{c}{N/A} \\
        & Ours & \textbf{0.439} & \textbf{0.684} & \textbf{0.916} & \textbf{0.882} & $\boldsymbol <$ \textbf{1}\\
        \bottomrule
    \end{tabular}
	\caption{Results on \shapesquare, \shapepentagon, and \shapehexagon~shapes. Since the number of vertices in each polygon for \shapepentagon~and \shapehexagon~is different, V-GAN is not applicable for those cases.\label{tab:exp_square_pentagon_hexagon}}
\end{table}

\begin{table}[t]
	\centering
	\scriptsize
	\begin{tabular}{cccccc}
		\toprule
		\textbf{Method} & \textbf{Cov@0.95} & \textbf{Cov@0.90} & \textbf{Cov} & \textbf{IoU} & \textbf{Time (sec.)} \\
		\midrule
		SA & 0.000 & 0.000 & 0.761 & 0.671 & 153 \\
		BayesOpt & 0.000 & 0.000 & 0.756 & 0.704 & 97 \\
		V-GAN & 0.000 & 0.000 & 0.549 & 0.545 & $\boldsymbol <$ \textbf{1} \\
		Ours & \textbf{0.384} & \textbf{0.545} & \textbf{0.892} & \textbf{0.854} & $\boldsymbol <$ \textbf{1} \\
		\bottomrule
	\end{tabular}
	\caption{Results on \shapemondrian~shape.\label{tab:exp_mondrian_square}}
\end{table}

To express a set of the fragment candidates to a latent representation $\bh$, 
we utilize the multi-head attention:
\begin{align}
    \calH &= \MH(\calX, \calX, \calX) \in \bbR^{|\calX| \times d}, \label{eqn:mh_1}\\
    \bh &= \MH(\mathbf{1}, \calH, \calH) \in \bbR^{d}, \label{eqn:mh_2}
\end{align}
where $\mathbf{1} \in \bbR^{1 \times d}$ is an all-ones matrix.
Without loss of generality, \eqnref{eqn:mh_1} and \eqnref{eqn:mh_2} can take $\calX$ (or $\calX'_i$)
and $\calH$, respectively.
Moreover, \eqnref{eqn:mh_1} can be stacked by feeding 
$\calH$ to the multi-head attention as an input, 
and \eqnref{eqn:mh_2} can be stacked in the similar manner.
Note that $\bh$ becomes $\bbR^{1 \times d}$, but we can employ it as 
a $\bbR^d$-shaped representation.
Additionally, FRAM includes a skip connection between an input and an output,
in order to partially hold the input information.

FRAM is a generalization of global feature aggregation 
such as average pooling and max pooling across instances, 
so that it considers high-order interaction between instances 
and aggregates intermediate instance-wise representations (i.e., $\calH$) 
with learnable parameters.
In \secref{sec:exp}, 
we show that this module is powerful for assembling fragments via elaborate studies on geometric shape assembly.

The final output is determined by selecting the fragment with 
the index of maximum $y_i$: $i^\star = \argmax_{i \in \{1, \ldots, |\calX|\}} y_i$.
After choosing the next fragment, its pose is selected through the maximum probability on $\bM_{i^\star}$ (and $\br_{i^\star}$).

\subsection{Training\label{sec:training}}

We train our neural network 
based on the objective that combines losses for two sub-tasks, 
fragment selection, i.e., \eqnref{eqn:loss_select} and fragment placement, i.e., \eqnref{eqn:loss_place}.
Since we first choose the next fragment, and then determine the position and angle of 
the selected fragment, 
the gradient updates by the summation of two losses should be applied.
Given a batch size $M$, 
the loss for the fragment selection part is 
\begin{equation}
    \calL_{\textrm{select}} = \sum_{m = 1}^M - \bt_m^\top \log \sigma(\by_m),
    \label{eqn:loss_select}
\end{equation}
where $\bt_m$ is one-hot encoding of true fragment, 
each entry of $\by_m$ is computed by \eqnref{eqn:frn_select}, 
and $\sigma$ is a softmax function.

Next, the objective for the fragment placement part is 
\begin{align}
    \calL_{\textrm{pose}} &= -\sum_{m = 1}^M \sum_{l = 0}^L \vectorize(\textrm{max-p}_{2l}(\bstau_m))^{\top} \log \vectorize(\textrm{avg-p}_{2l}(\bM_m))\nonumber\\
    &\quad- \sum_{m = 1}^M {\boldsymbol \rho}_m^\top \log \sigma(\br_m),
    \label{eqn:loss_place}
\end{align}
where $\bstau_m \in \bbR^{w \times h}$/${\boldsymbol \rho}_m \in \bbR^{r}$ are one-hot encodings 
of true position/true angle, 
$\bM_m$ (and $\br_m$) are computed by \eqnref{eqn:frn_place}, 
$L$ is the number of pooling operations, 
and $\textrm{max-p}$/$\textrm{avg-p}$/$\vectorize$ are functions for applying max-pooling/applying average-pooling/vectorizing a matrix.
The subscript of pooling operations indicates the size of pooling and stride, and zero-sized pooling returns an original input.
Moreover, the coefficients for balancing the objective functions should be multiplied to each term of the objectives; see the supplementary material.
% Experimental Results
% !TEX root =  ../main.tex

\section{Experimental Results\label{sec:exp}}

We show that the qualitative and quantitative results comparing to the three baselines for four target geometric shapes. We also provide studies on interesting assembly scenarios.

\paragraph{Details of Experiments.}

As shown in \figref{fig:partition}, 
we evaluate our method and existing methods on the following target objects:
\shapesquare, \shapemondrian, \shapepentagon, and \shapehexagon.
Unless specified otherwise, we use 5,000 samples each of them is partitioned into 8 fragments using binary space partitioning and the number of rotation angle bins are set to 1, 4, or 20.
We use 64\%, 16\%, and 20\% of the samples for training, validation, and test splits, respectively.
An assembly quality is calculated with coverage (Cov) and Intersection over Union (IoU).
\begin{equation}
    \textrm{Cov}(\bc, S) = \frac{\textrm{area}(\bc \cap S)}{\textrm{area}(S)} \ \
    \textrm{and} \ \
    \textrm{IoU}(\bc, S) = \frac{\textrm{area}(\bc \cap S)}{\textrm{area}(\bc \cup S)},
\end{equation}
where $\bc$ is an assembled shape and $S$ is a target shape.

\begin{figure}[t]
    \centering
        \subfigure[Simulated annealing]{
            \includegraphics[width=0.11\textwidth]{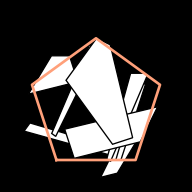}
            \includegraphics[width=0.11\textwidth]{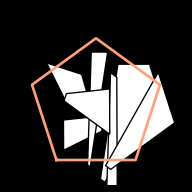}
            \includegraphics[width=0.11\textwidth]{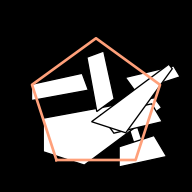}
            \includegraphics[width=0.11\textwidth]{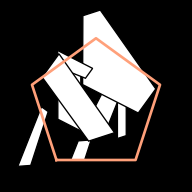}
        }
        \subfigure[Bayesian optimization]{
            \includegraphics[width=0.11\textwidth]{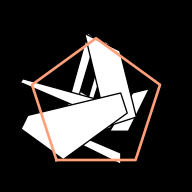}
            \includegraphics[width=0.11\textwidth]{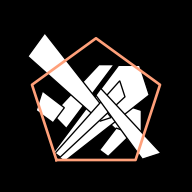}
            \includegraphics[width=0.11\textwidth]{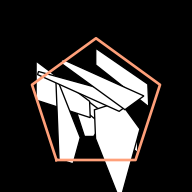}
            \includegraphics[width=0.11\textwidth]{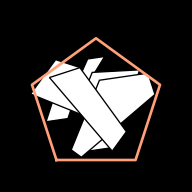}
        }
        \subfigure[Ours]{
            \includegraphics[width=0.11\textwidth]{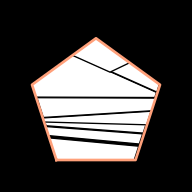}
            \includegraphics[width=0.11\textwidth]{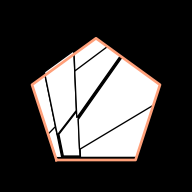}
            \includegraphics[width=0.11\textwidth]{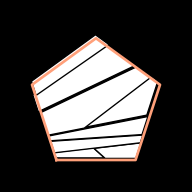}
            \includegraphics[width=0.11\textwidth]{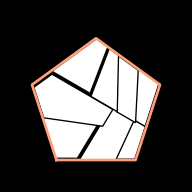}
        }
        \caption{Assembling results for \shapepentagon.\label{fig:assembly_pentagon}}
\end{figure}

\begin{figure}[t]
        \centering
        \subfigure[Simulated annealing]{
            \includegraphics[width=0.11\textwidth]{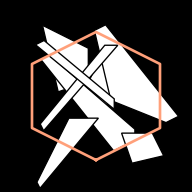}
            \includegraphics[width=0.11\textwidth]{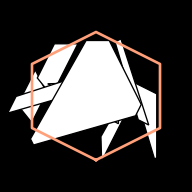}
            \includegraphics[width=0.11\textwidth]{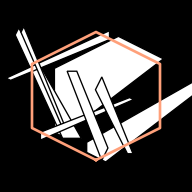}
            \includegraphics[width=0.11\textwidth]{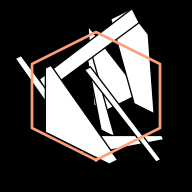}
        }
        \subfigure[Bayesian optimization]{
            \includegraphics[width=0.11\textwidth]{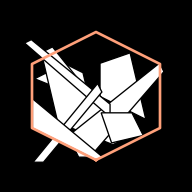}
            \includegraphics[width=0.11\textwidth]{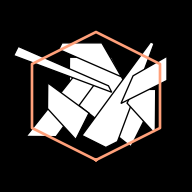}
            \includegraphics[width=0.11\textwidth]{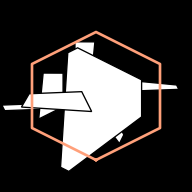}
            \includegraphics[width=0.11\textwidth]{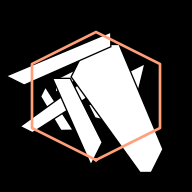}
        }
        \subfigure[Ours]{
            \includegraphics[width=0.11\textwidth]{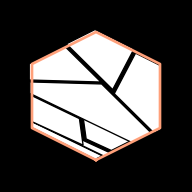}
            \includegraphics[width=0.11\textwidth]{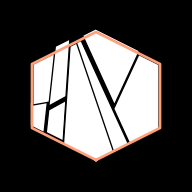}
            \includegraphics[width=0.11\textwidth]{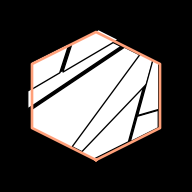}
            \includegraphics[width=0.11\textwidth]{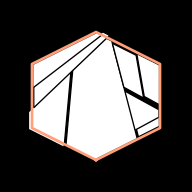}
        }
        \caption{Assembling results for \shapehexagon.\label{fig:assembly_hexagon}}
\end{figure}

Since there exist no prior methods that specifically aim to tackle shape assembly of textureless fragments with indistinctive junctions, we compare ours with two classic optimization methods, simulated annealing~\cite{PincusM1970or}, Bayesian optimization~\cite{BrochuE2010arxiv}, and a generative method using generative adversarial networks~\cite{LiJ2020tpami}.
The details can be found in the supplementary material.

\subsection{Shape Assembly Experiments}

Experiments in this section compare our approach to other baseline methods on the assembly of different target objects.
The quantitative results on \shapesquare, \shapemondrian, \shapepentagon, and \shapehexagon~objects are summarized in \tabref{tab:exp_square_pentagon_hexagon}. 
We find that our method consistently covers much more area than baselines within less time budget.

The qualitative results with randomly partitioned fragments from geometric shapes such as \shapesquare, \shapepentagon, and \shapehexagon~are presented in \figref{fig:assembly_square}, \figref{fig:assembly_pentagon}, and \figref{fig:assembly_hexagon}.
In \figref{fig:assembly_mondrian}, the results with axis-aligned partitioning process, \shapemondrian, is also presented.

V-GAN requires a fixed number of vertices for partitioned fragments, which is only applicable to \shapesquare~and \shapemondrian~cases.
Both SA and BayesOpt struggle to place fragments in the correct position, 
but many overlaps occur between fragments.
They also consume longer time than our method in all experimental conditions.
On the other hand, V-GAN consumes less time than other baseline methods while it significantly underperforms compared to other methods.

\begin{figure*}[t]
	\centering
	\includegraphics[width=0.95\textwidth]{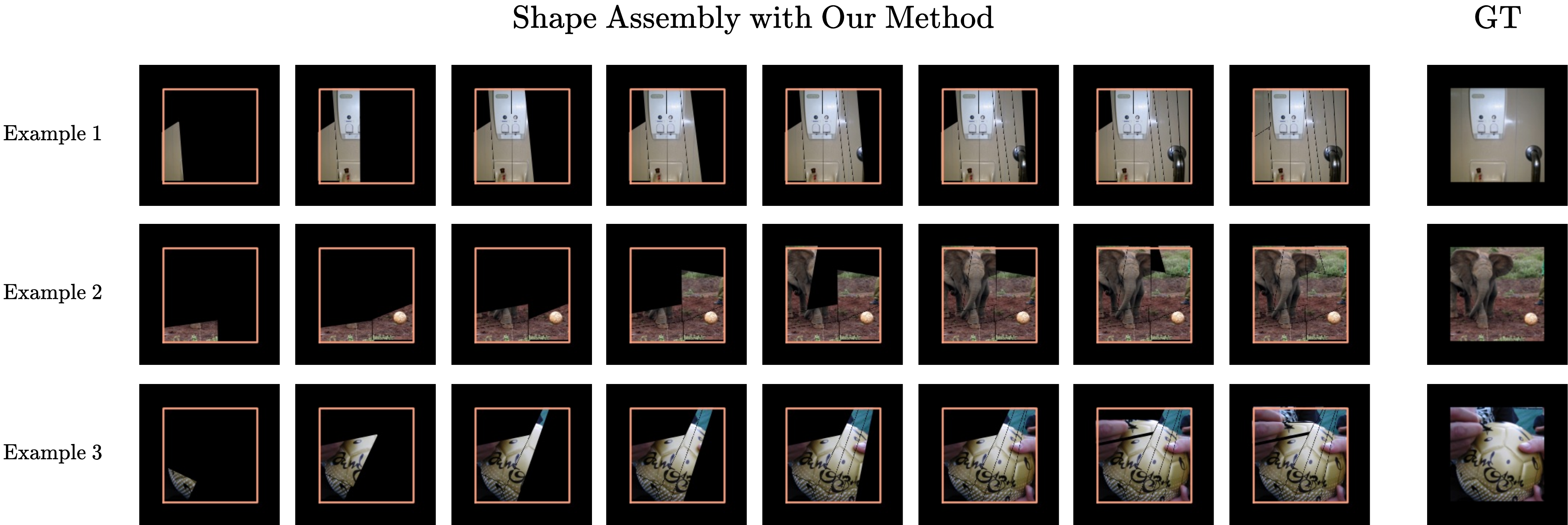}
	\caption{Assembling results for ImageNet examples with colored textures.\label{fig:texture}}
\end{figure*}

\subsection{Elaborate Study \& Ablation Study}
We present interesting studies on \texttt{Mondrian-Square} objects, the number of fragments, abnormal fragments, rotation discretization, the effectiveness of FRAM, sequence sampling strategy, and scenarios with texture.

\begin{figure}[t]
	\centering
	\subfigure[Missing]{
		\includegraphics[width=0.11\textwidth]{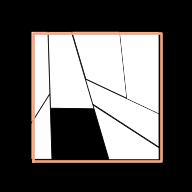}
	}
	\subfigure[Eroded]{
		\includegraphics[width=0.11\textwidth]{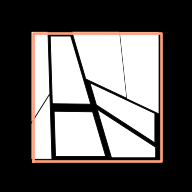}
	}
	\subfigure[Distorted]{
		\includegraphics[width=0.11\textwidth]{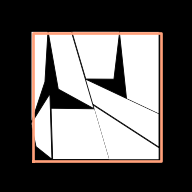}
	}
	\caption{Scenarios with abnormal fragments.\label{fig:different_scenarios}}
\end{figure}

\begin{table}[t]
	\centering
	\scriptsize
	\setlength{\tabcolsep}{5pt}
	\begin{tabular}{cccccc}
		\toprule
		\textbf{\#Frags} & \textbf{Target Shape} & \textbf{Cov@0.95} & \textbf{Cov@0.90} & \textbf{Cov} & \textbf{IoU} \\
		\midrule
		\multirow{3}{*}{4 fragments} & \texttt{Square} & 0.972 & 0.989 & 0.991 & 0.962 \\
		& \texttt{Pentagon} & 0.962 & 0.974 & 0.991 & 0.958 \\
		& \texttt{Hexagon} & 0.940 & 0.976 & 0.989 & 0.956 \\
		\midrule
		\multirow{3}{*}{16 fragments} & \texttt{Square} & 0.010 & 0.076 & 0.769 & 0.730 \\
		& \texttt{Pentagon} & 0.024 & 0.108 & 0.797 & 0.755 \\
		& \texttt{Hexagon} & 0.008 & 0.062 & 0.771 & 0.729 \\
		\bottomrule
	\end{tabular}
	\caption{Study on shape assemblies with 4 and 16 fragments.\label{tab:ablation_number_frag}}
\end{table}

\begin{table}[t]
	\centering
	\scriptsize
	\setlength{\tabcolsep}{4pt}
	\begin{tabular}{cccccc}
		\toprule
		\textbf{Target Shape} & \textbf{Abnormal Type} & \textbf{Cov@0.95} & \textbf{Cov@0.90} & \textbf{Cov} & \textbf{IoU} \\
		\midrule
		\multirow{3}{*}{\texttt{Square}} & Missing & 0.125 & 0.254 & 0.808 & 0.716 \\
		& Eroded & 0.108 & 0.322 & 0.843 & 0.803 \\
		& Distorted & 0.010 & 0.060 & 0.753 & 0.685 \\
		\midrule
		\multirow{3}{*}{\texttt{Pentagon}} & Missing & 0.208 & 0.407 & 0.852 & 0.764 \\
		& Eroded & 0.049 & 0.197 & 0.829 & 0.791 \\
		& Distorted & 0.031 & 0.113 & 0.793 & 0.739 \\
		\midrule
		\multirow{3}{*}{\texttt{Hexagon}} & Missing & 0.253 & 0.459 & 0.855 & 0.778 \\
		& Eroded & 0.065 & 0.221 & 0.814 & 0.773 \\
		& Distorted & 0.026 & 0.125 & 0.778 & 0.723 \\
		\bottomrule
	\end{tabular}
	\caption{Study on missing, eroded, and distorted scenarios.\label{tab:exp_missing_eroded_distorted}}
\end{table}

\begin{table}[t]
	\centering
	\scriptsize
	\begin{tabular}{cccccc}
		\toprule
		\textbf{Target Shape} & \textbf{\#Bins} & \textbf{Cov@0.95} & \textbf{Cov@0.90} & \textbf{Cov} & \textbf{IoU} \\
		\midrule
		\multirow{2}{*}{\texttt{Square}} & 4 & 0.010 & 0.017 & 0.670 & 0.622 \\
		& 20 & 0.000 & 0.000 & 0.603 & 0.519 \\
		\midrule
		\multirow{2}{*}{\texttt{Pentagon}} & 4 & 0.062 & 0.147 & 0.791 & 0.726 \\
		& 20 & 0.000 & 0.000 & 0.604 & 0.515 \\
		\midrule
		\multirow{2}{*}{\texttt{Hexagon}} & 4 & 0.009 & 0.046 & 0.731 & 0.676 \\
		& 20 & 0.000 & 0.000 & 0.640 & 0.550 \\
		\bottomrule
	\end{tabular}
	\caption{Study on rotation discretization.\label{tab:ablation_rotation}}
\end{table}

\begin{table}[t]
	\centering
	\scriptsize
	\begin{tabular}{ccccc}
		\toprule
		\textbf{Method} & \textbf{Cov@0.95} & \textbf{Cov@0.90} & \textbf{Cov} & \textbf{IoU} \\
		\midrule
		w/o Aggregation & 0.392 & 0.569 & 0.899 & 0.863 \\
		Max Pooling & 0.390 & 0.563 & 0.893 & 0.856 \\
		Avg Pooling & 0.431 & 0.580 & 0.899 & 0.863 \\
		w/ FRAM & \textbf{0.470} & \textbf{0.649} & \textbf{0.914} & \textbf{0.882} \\
		\bottomrule
	\end{tabular}
	\caption{Study on feature aggregation methods.\label{tab:ablation_number_fram}}
\end{table}

\begin{table}[t]
	\centering
	\scriptsize
	\begin{tabular}{cccccc}
		\toprule
		\textbf{Target Shape} & \textbf{Strategy} & \textbf{Cov@0.95} & \textbf{Cov@0.90} & \textbf{Cov} & \textbf{IoU} \\
		\midrule
		\multirow{2}{*}{\texttt{Square}} & Random & 0.012 & 0.109 & 0.814 & 0.770 \\
		& $\nearrow$ & 0.470 & 0.649 & 0.914 & 0.882 \\
		\midrule
		\multirow{2}{*}{\texttt{Pentagon}} & Random & 0.016 & 0.115 & 0.802 & 0.751 \\
		& $\nearrow$ & 0.452 & 0.696 & 0.922 & 0.884 \\
		\midrule
		\multirow{2}{*}{\texttt{Hexagon}} & Random & 0.034 & 0.146 & 0.807 & 0.762 \\
		& $\nearrow$ & 0.439 & 0.684 & 0.916 & 0.882 \\
		\bottomrule
	\end{tabular}
	\caption{Study on sequence sampling strategy. $\nearrow$ indicates from bottom, left to top, right.\label{tab:ablation_simpler}}
\end{table}

\paragraph{Axis-Aligned Space Partitioning.}

To create a more challenging problem, 
we apply axis-aligned space partitioning in the shape fragmentation process.
As shown in~\figref{fig:partition}, 
it generates similar rectangle fragments, 
which can puzzle our method.
Nevertheless, our network successfully 
assembles the fragments given, as shown in~\figref{fig:assembly_mondrian} and \tabref{tab:exp_mondrian_square}.

\paragraph{Number of Fragments.}
As mentioned before, our model can be applied to the dataset 
that consists of a different number of fragments.
Thus, we prepare a dataset, 
given the number of partitions $K = 2$ or $K = 4$.
The results of our proposed model on the datasets with 4 fragments and 16 fragments 
of three geometric shapes are shown in \tabref{tab:ablation_number_frag}.
If the number of fragments is 4, 
it performs better than the circumstance with 8 fragments.
On the contrary, a dataset with 16 fragments 
deteriorates the performance, 
but our method covers the target object with relatively small losses.
    
\paragraph{Abnormal Fragments.}
We consider a particular situation to 
reconstruct a target object with missing, eroded, or distorted fragments, 
which can be thought of as the excavation scenario in archaeology;
see \figref{fig:different_scenarios}.
We assume one fragment is missing, 
or four fragments are eroded or distorted randomly.
We measure the performance of our method 
excluding the degradation from the ground-truth target object.
Our methods outperform other methods 
as shown in \tabref{tab:exp_missing_eroded_distorted}.

\paragraph{Rotation Discretization.}
To show the effects of rotation discretization, 
we test two bin sizes.
As shown in~\tabref{tab:ablation_rotation}, 
delicate discretization is more difficult than 
the case with a small bin size, which follows our expectation.

\paragraph{Effectiveness of FRAM.}
It empirically validates the effectiveness of FRAM.
As summarized in \tabref{tab:ablation_number_fram}, 
FRAM outperforms other aggregation methods.
The w/o aggregation model does not have a network for relations between fragments, 
and the other two models replace our FRAM to the associated aggregation methods among feature vectors.

\paragraph{Sequence Sampling Strategy.} 
As shown in \tabref{tab:ablation_simpler}, 
our sampling strategy that samples from bottom, left to top, right is 
effective compared to the random sampling technique that 
chooses the order of fragments based on the proximity of the fragments previously sampled.

\paragraph{Scenarios with Texture.} 
Our model is also able to handle a scenario with colored textures by allowing it to take 
RGB images.
As shown in~\figref{fig:texture}, 
our method successfully assembles 
fragments with texture into the target object.
% Discussion
% !TEX root =  ../main.tex

\section{Discussion \& Related Work\label{sec:discussion}}

Compared to generic setting of jigsaw puzzle solvers~\cite{ChoTS2010cvpr,NorooziM2016eccv}, 
our problem is more challenging as 
it involves textureless fragments and indistinctive junctions 
as well as arbitrary shapes and rotatable fragments.
As presented in \tabref{tab:exp_square_pentagon_hexagon}, 
the empirical results for \shapepentagon~and \shapehexagon~tend 
to be better than the results for \shapesquare.
It implies that if a target shape becomes complicated and contains more information, 
the problem becomes more easily solvable.
This tendency is also shown in the experiments with textures; see~\figref{fig:texture}.

The results with axis-aligned partitioning process, \shapemondrian, is also presented in \tabref{tab:exp_mondrian_square}.
The outcome by this process, which resembles the paintings by Piet Mondrian as shown in the second example of \figref{fig:partition}, 
can be more difficult than the randomly partitioned shape,
because fragment shapes and junction information are less expressive than arbitrary fragments.
However the results show that our method is still more effective than the other baselines despite the performance loss, 
which implies that our method is able to select and place fragments 
although ambiguous choices exist; see~\figref{fig:assembly_mondrian}.

Unlike our random fragmentation, the approaches to solving a sequential assembly problem with fixed primitives~\cite{BapstV2019icml,KimJ2020ml4eng,ChungH2021neurips} have been studied recently.
However, they focus on constructing a physical object by assembling given unit primitives,
and they utilize explicit supervision such as volumetric (or areal) comparisons and particular rewards.
% Conclusion
% !TEX root =  ../main.tex

\section{Conclusion\label{sec:conclusion}}
In this paper, we have solved a two-dimensional geometric shape assembly problem.
Our model, dubbed FAN, predicts the next fragment and its pose where fragments to assemble are given, 
with an attention-based module, dubbed FRAM.
We showed that our method outperforms other baseline methods 
including simulated annealing, Bayesian optimization, and a learning-based adversarial network.
Moreover, we evaluated our methods in diverse circumstances, 
such as novel assembly scenarios with axis-aligned space partitioning, 
degraded fragments, and colored textures.

\section*{Acknowledgments}

This work was supported by Samsung Research Funding \& Incubation Center of Samsung Electronics under Project Number SRFC-TF2103-02.

\bibliographystyle{named}
\bibliography{ijcai22}

\clearpage

\renewcommand{\theequation}{s.\arabic{equation}}
\renewcommand{\thetable}{s.\arabic{table}}
\renewcommand{\thefigure}{s.\arabic{figure}}
\renewcommand{\thealgorithm}{s.\arabic{algorithm}}
\renewcommand{\thesection}{S.\arabic{section}}
\renewcommand{\thesubsection}{S.\arabic{section}.\arabic{subsection}}

\setcounter{section}{0}
\setcounter{algorithm}{0}
\setcounter{equation}{0}
\setcounter{figure}{0}
\setcounter{table}{0}

\noindent {\Large \textbf{Supplementary Material}}
\vspace{20pt}

In this material, we describe the detailed contents 
that supplement our main article.

\section{Details of Shape Assembly\label{sec:suppl_assembly}}

\begin{algorithm}[ht]
	\caption{Geometric Shape Assembly Procedure}
	\label{alg:assembly}
	\begin{algorithmic}[1]
		\REQUIRE Fragments $\calX = \{ \bx_i \}_{i=1}^{N}$ and a target shape $S$.
		\ENSURE Output shape $\bc$.
		\STATE Initialize a current shape $\bc$ on a space $\Phi$.
		\WHILE { $\calX \neq \emptyset$}
            \STATE Select $\bx_i \in \calX$ and place it on $\Phi$.
            \STATE Assemble $\bc \leftarrow$ shape $\bc$ updated by the placement of $\bx_i$.
		    \STATE Update $\calX \leftarrow \calX \setminus \bx_i$.
        \ENDWHILE
		\STATE \textbf{return} $\bc$
	\end{algorithmic}
\end{algorithm}

We provide the detailed procedure of shape assembly in~\algref{alg:assembly}.

\section{Details of Shape Fragmentation\label{sec:suppl_frag}}

We present the procedure of shape fragmentation 
in~\algref{alg:partition}.

\begin{algorithm}[ht]
	\caption{Shape Fragmentation}
	\label{alg:partition}
	\begin{algorithmic}[1]
		\REQUIRE A target shape $S$, 
		    the number of partitions $K$.
		\ENSURE Fragments partitioned $\calX_K = \{ \bx_i \}_{i = 1}^{2^K}$.
		\STATE Initialize a set of fragments $\calX_0 = \{ S \}$.
		\FOR {$i =  1, \ldots, K$}
		    \STATE Initialize $\calX_i = \{ \}$.
		    \FORALL {$\bx \in \calX_{i - 1}$}
		        \STATE Partition a fragment $\bx$ to $\bx_{+}, \bx_{-}$.
		        \STATE Update $\calX_i \leftarrow \calX_i + \{ \bx_{+}, \bx_{-} \}$.
		    \ENDFOR
		\ENDFOR
		\STATE Rotate all the fragments in $\calX_K$ at random.
		\STATE \textbf{return} a set of fragments $\calX_K$
	\end{algorithmic}
\end{algorithm}

\begin{figure}[ht]
    \centering
    \subfigure[Partitioning 1]{
        \includegraphics[width=0.14\textwidth]{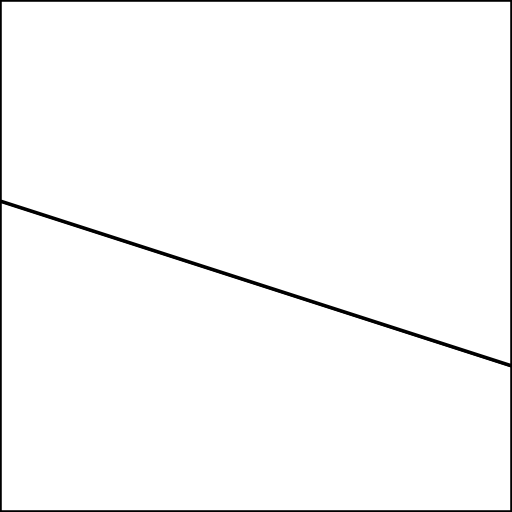}
    }
    \subfigure[Partitioning 2]{
        \includegraphics[width=0.14\textwidth]{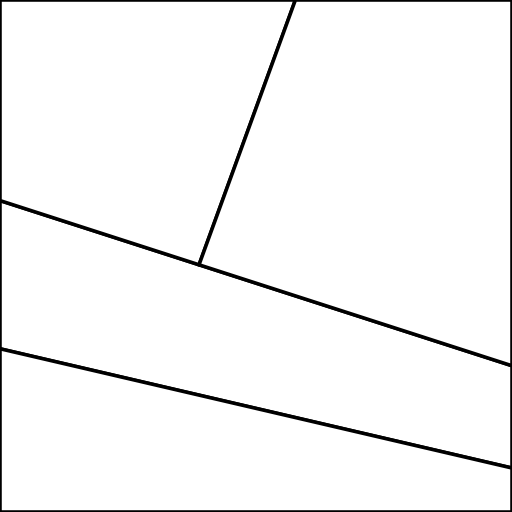}
    }
    \subfigure[Partitioning 3]{
        \includegraphics[width=0.14\textwidth]{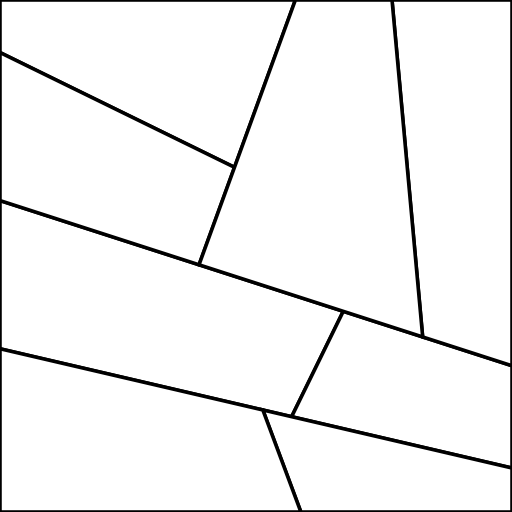}
    }
    \caption{Fragmentation examples on \shapesquare~by the number of partitions.}
    \label{fig:partitioning}
\end{figure}

We divide a target geometric shape into fragments using a binary space partitioning algorithm~\cite{SchumacherR1969techreport}.
The number of fragments increases exponentially with the number of partitions.
Some of partitioning examples are shown in~\figref{fig:partitioning}.

We create a shape fragmentation dataset under 
two constraints that prevent generating small fragments.
First, the hyperplane does not cross adjacent edges of the given polygon.
Second, the position that the hyperplane will pass through 
is randomly selected between the range of 
$\pm$ 25$\%$ of edge length from the center of the edge.
The dataset we use is composed of 5,000 samples.

\paragraph{Sequence Sampling Strategy.}
As presented in~\tabref{tab:ablation_simpler}, 
the consistency of assembling sequences in the dataset turns out to be crucial in training our network.
If we employ the assembly order randomly determined in training our network, 
it performs significantly worse.
To achieve better performance, 
we should select the next fragment that is adjacent to the fragments already chosen.
As inspired by human assembling process~\cite{ChazelleB1983ieeetc}, 
we determine the order of fragments from bottom, left to top, right 
by considering the center position of fragments in the target shape.

\section{Details of Experiments\label{sec:suppl_exp}}

Our framework is implemented using TensorFlow~\cite{AbadiM2016osdi}.
We conduct the experiments on a Ubuntu 18.04 workstation
with twelve Intel(R) Core i7-5930K CPU running at 3.50 GHz,
64 GB random access memory and NVIDIA Titan X (pascal) GPU.

To train our model with the objectives defined in the main article, 
we need to balance the terms in two objectives.
Thus, we do our best to find the coefficients for these objectives.
As a result, 
we scale up the term for determining the position of the next fragment 
by multiplying 1,000 and the term for a rotation probability 
by multiplying 10.

\subsection{Baselines}

We describe the details of baselines we employ in this paper.

\paragraph{Simulated annealing (SA).}
Similar to our proposed method, SA~\cite{PincusM1970or} iteratively selects and places one fragment such that IoU between the target object and assembled fragments is maximized.
Thus, the optimization is repeated until the number of assembled fragments is equal to the given total budget.

\paragraph{Bayesian optimization (BayesOpt).}
BayesOpt~\cite{BrochuE2010arxiv} is used to find the pose of given fragments by maximizing IoU between fragments and a target object.
Similar to SA, the optimization is repeated in total of the number of fragments.
We utilize the Bayesian optimization package~\cite{KimJ2017bayeso} to implement this baseline.

\paragraph{LayoutGAN modified for vertex inputs (V-GAN).}
It is a modification of \cite{LiJ2020tpami}.
Since LayoutGAN assumes that fragment shapes are always fixed 
and only finds their positions, we modify the generator structure 
to a neural network that takes fragment vertices as inputs and predicts center coordinates as their positions.
Discriminator then makes real or fake decision
based on adjusted vertex coordinates.
Similar to GAN~\cite{GoodfellowIJ2014neurips}, 
it is trained via an adversarial training strategy.

\subsection{Model Architecture}

We introduce the model architecture for FAN including FRAM and V-GAN.

\begin{table}[t]
	\centering
    \begin{tabular}{ccc}
        \toprule
        \textbf{Network} & \textbf{Layer} & \textbf{Output Dimension} \\
        \midrule
        Generator & Reshape &  64 \\
        & FC & 128 \\
        & BatchNorm & 128 \\
        & ReLU & 128 \\
        & FC & 64 \\
        & BatchNorm & 64 \\
        & ReLU & 64 \\
        & FC & 32 \\
        & BatchNorm & 32 \\
        & ReLU & 32 \\
        & FC & 16 \\
        \midrule
        Discriminator & Reshape & 64 \\
        & FC & 128 \\
        & ReLU & 128 \\
        & Dropout 0.3 & 128 \\
        & FC & 64 \\
        & ReLU & 64 \\
        & Dropout 0.3 & 64 \\
        & FC & 32 \\
        & ReLU & 32 \\
        & Dropout 0.3 & 32 \\
        & FC & 16 \\
        \bottomrule
    \end{tabular}
	\caption{Architecture of V-GAN.\label{tab:suppl_arch_2}}
\end{table}

\begin{table*}[p]
	\centering
    \begin{tabular}{ccc}
        \toprule
        \textbf{Network} & \textbf{Layer} & \textbf{Output Dimension} \\
        \midrule
        Encoder & Conv 2D, 16 ch., 3 $\times$ 3, stride 1, padding 1 & 128 $\times$ 128 $\times$ 16 \\
        & ReLU & 128 $\times$ 128 $\times$ 16 \\
        & Dropout 0.1 & 128 $\times$ 128 $\times$ 16 \\
        & Conv 2D, 16 ch., 3 $\times$ 3, stride 1, padding 1 & 128 $\times$ 128 $\times$ 16 \\
        & ReLU & 128 $\times$ 128 $\times$ 16 \\
        & Maxpool 2D, 2 $\times$ 2, stride 2, padding 0 & 64 $\times$ 64 $\times$ 16 \\
        & Conv 2D, 32 ch., 3 $\times$ 3, stride 1, padding 1 & 64 $\times$ 64 $\times$ 32 \\
        & ReLU & 64 $\times$ 64 $\times$ 32 \\
        & Dropout 0.1 & 64 $\times$ 64 $\times$ 32 \\
        & Conv 2D, 32 ch., 3 $\times$ 3, stride 1, padding 1 & 64 $\times$ 64 $\times$ 32 \\
        & ReLU & 64 $\times$ 64 $\times$ 32 \\
        & Maxpool 2D, 2 $\times$ 2, stride 2, padding 0 & 32 $\times$ 32 $\times$ 32 \\
        & Conv 2D, 64 ch., 3 $\times$ 3, stride 1, padding 1 & 32 $\times$ 32 $\times$ 64 \\
        & ReLU & 32 $\times$ 32 $\times$ 64 \\
        & Dropout 0.2 & 64 $\times$ 64 $\times$ 32 \\
        & Conv 2D, 64 ch., 3 $\times$ 3, stride 1, padding 1 & 32 $\times$ 32 $\times$ 64 \\
        & ReLU & 32 $\times$ 32 $\times$ 64 \\
        & Flatten & 65,536 \\
        & Linear & 256 \\
        & ReLU & 256 \\
        \midrule
        Decoder & Linear & 32,768  \\
        & ReLU & 32,768 \\
        & Reshape & 32 $\times$ 32 $\times$ 32 \\
        & ConvTranspose 2D, 32 ch., 2 $\times$ 2, stride 2, padding 2 & 64 $\times$ 64 $\times$ 32 \\
        & Conv 2D, 32 ch., 3 $\times$ 3, stride 1, padding 1 & 64 $\times$ 64 $\times$ 32 \\
        & ReLU & 64 $\times$ 64 $\times$ 32 \\
        & Dropout 0.1 & 64 $\times$ 64 $\times$ 32 \\
        & Conv 2D, 32 ch., 3 $\times$ 3, stride 1, padding 1 & 64 $\times$ 64 $\times$ 32 \\
        & ReLU & 64 $\times$ 64 $\times$ 32 \\
        & ConvTranspose 2D, 16 ch., 2 $\times$ 2, stride 2, padding 2 & 128 $\times$ 128 $\times$ 32 \\
        & Conv 2D, 32 ch., 3 $\times$ 3, stride 1, padding 1 & 128 $\times$ 128 $\times$ 16 \\
        & ReLU & 128 $\times$ 128 $\times$ 16 \\
        & Dropout 0.1 & 128 $\times$ 128 $\times$ 16 \\
        & Conv 2D, 32 ch., 3 $\times$ 3, stride 1, padding 1 & 128 $\times$ 128 $\times$ 16 \\
        & ReLU & 128 $\times$ 128 $\times$ 16 \\
        & Conv 2D, 1 ch., 1 $\times$ 1, stride 1, padding 0 & 128 $\times$ 128 $\times$ 1 \\
        \midrule
        MLP & Linear & 256 \\
        & Batch normalization & 256 \\
        & ReLU & 256 \\
        & Linear & 256 \\
        & Batch normalization & 256 \\
        & ReLU & 256 \\
        & Linear & 1 or 256 \\
        \midrule
        FRAM & &  \\
        FRAM-Encoder & Encoder layer $\times$ 2 (including MHA. $\times$ 8) & 256 \\
        FRAM-Decoder & Decoder layer $\times$ 2 (including MHA. $\times$ 8) & 256 \\
        \bottomrule
    \end{tabular}
	\caption{Architecture of FAN.\label{tab:suppl_arch_1}}
\end{table*}

\paragraph{FAN.}
FAN consists of $\frnsel$ that selects next fragment and $\frnloc$ 
that predicts center coordinates of the chosen fragment.
There exist 7 hyperparameters in FAN with 3 specifically in FRAM.
Full details of the pipeline can be found in \tabref{tab:suppl_arch_1}.
The default dimension of fully connected layer in the whole pipeline 
is fixed to 256 throughout the experiments.
For the training, we use batch size of 32 and Adam optimizer~\cite{KingmaDP2015iclr} 
for both $\frnsel$ and $\frnloc$ 
with learning rate of $1 \times 10^{-3}$ respectively.
$\frnsel$ is a convolutional neural network (CNN) followed by FRAM including MLP block.
$\frnloc$ is convolutional encoder-decoder with MLP block followed by an additional fully-connected layer inserted in between. Each encoder for $\frnloc$ and $\frnsel$ do not share the weights, and thus are trained with different loss. Our FRAM follows a similar setting of the original Transformer model~\cite{VaswaniA2017neurips} with 2,048 hidden units and 8 attention heads.

\paragraph{V-GAN.}
This model follows typical setup of generative adversarial network~\cite{GoodfellowIJ2014neurips}, having both generator and discriminator.
Specifically, a generator takes fragment’s vertex coordinates as inputs and predicts center coordinates.
Then fragment's vertex coordinates are translated by predicted center coordinates.
Next, a discriminator makes fake or real decision based on translated  vertex coordinates.
For training, we use batch size of 128 and Adam optimizer with learning rate of $1 \times 10^{-4}$. The details of model is summarized in \tabref{tab:suppl_arch_2}.

\section{Additional Discussion\label{sec:suppl_discussion}}

Our proposed network can be viewed as a permutation-equivariant network, which satisfies
$f(\pi([\bx_1, \ldots, \bx_n]^\top)) = \pi[f([\bx_1, \ldots, \bx_n]^\top)]$,
where $f: \bbR^{n \times d} \to \bbR^{n \times d'}$ is a neural network that outputs a set and $\pi$ is a permutation function along the first dimension.
The work~\cite{GuttenbergN2016arxiv} suggests a permutational layer to handle the different inputs and does not depend on the specific order of inputs.
The work of~\cite{ZaheerM2017neurips} derives the necessary and sufficient conditions on permutation-equivariant networks.
Moreover, an attention-based block for sets has been proposed~\cite{LeeJ2019icml}.

Furthermore, the number of learnable parameters does not depend on 
the dimensionality of $\by$ and the sizes of $\bM_i$ and $\br_i$ for $i \in \{1, \ldots, |\calX|\}$.
As shown in \eqnref{eqn:frn_select} and \eqnref{eqn:frn_place}, 
their outputs can handle variable-sized $\calX$ 
and the order of $\calX$ affects the order of the outputs, 
satisfying the permutation equivariance.
The experimental results support that our method can learn 
these challenging scenarios.

\end{document}